\documentclass[sigconf,nonacm]{acmart}

\usepackage{tikz}
\usetikzlibrary{arrows.meta,positioning,fit,backgrounds}
\usepackage{pgfplots}
\usepackage{enumitem}
\usepackage{fancyvrb}
\pgfplotsset{compat=1.18}

\newcommand{\benchmark}{\textsc{AgentChaosBench}}
\newcommand{\faultname}[1]{\textsf{#1}}
\newcommand{\stitle}[1]{\noindent{\textbf{#1.}}}

\title{When Agentic Executions Fail: Detecting and Localizing Runtime Faults from Telemetry}

\author{Chenkai Zhang}
\authornote{Both authors contributed equally to this research.}
\affiliation{%
  \institution{University of Toronto}
  \city{Toronto}
  \country{Canada}
}
\email{zhangchenkai.zhang@mail.utoronto.ca}

\author{Yiran Li}
\authornotemark[1]
\affiliation{%
  \institution{University of Toronto}
  \city{Toronto}
  \country{Canada}
}
\email{one.li@utoronto.ca}

\author{Yifang Tian}
\affiliation{%
  \institution{University of Toronto}
  \city{Toronto}
  \country{Canada}
}
\email{yifang.tian@mail.utoronto.ca}

\author{Michalis Bachras}
\affiliation{%
  \institution{University of Toronto}
  \city{Toronto}
  \country{Canada}
}
\email{michalis.bachras@mail.utoronto.ca}

\author{Hans-Arno Jacobsen}
\affiliation{%
  \institution{University of Toronto}
  \city{Toronto}
  \country{Canada}
}
\email{jacobsen@eecg.toronto.edu}

\begin{document}

\begin{abstract}
Reliability in LLM-based agentic systems is a property of the whole execution (its
tool calls, model calls, guardrails, and inter-agent messages), not of the final
answer alone, yet evaluating only task outcomes reveals little about how or why a
run fails. We present \benchmark{}, a benchmark for detecting and localizing runtime
faults in agentic systems from their execution telemetry. We run five heterogeneous
applications that coordinate agents over the Agent-to-Agent protocol and call tools
through the Model Context Protocol, and inject ten types of operational fault
(unavailable or slow tools, corrupted or oversized responses, and delayed, looped,
or misrouted delegations and bypassed guardrails) at their tool, model, guardrail,
and inter-agent boundaries, alongside a no-fault control. The resulting dataset
contains $275$ sanitized traces: $250$ faulty executions spanning ten fault types
and $25$ no-fault controls. Each faulty trace is aligned with the no-fault execution
of the same input; fault-type labels and, where applicable, location labels are held
out from diagnosis. On structured single-trace inputs, a first set of zero-shot LLM
baselines shows the task is far from solved: local detectors up to $14$B parameters
reach only
$13.6$--$19.2\%$ top-1 fault-type accuracy and the frontier DeepSeek-v4-pro only
$24.8\%$, while jointly identifying the fault type and its location tops out at
$22\%$; reference-dependent faults (above all a bypassed guardrail) stay near-unsolved
from a single trace. An aligned reference improves selected relative faults but
does not resolve guardrail bypass. The held-out labels and compact
prediction format support reproducible comparison of LLM-based and non-LLM diagnosis
methods.
\end{abstract}

\maketitle

\section{Introduction}

Large language model (LLM) agents are evolving from conversational interfaces into software systems that plan, invoke tools, maintain state, and coordinate with other agents~\cite{he2025llm,zhao2026debugging}. Benchmarks such as AgentBench demonstrate the breadth of tasks that can be addressed through multi-step interaction with external environments~\cite{liu2024agentbench}. More recent AI-native systems further combine agents with standardized tool and communication protocols, including the Model Context Protocol (MCP) and Agent-to-Agent (A2A) communication, making system behavior depend on orchestration logic and networked services in addition to the underlying model~\cite{wang2026ainativebench}. As these dependencies grow, evaluating only the final task outcome provides little insight into how reliably the system operates or why a particular execution fails.

\begin{figure*}[t]
\centering
\begin{tikzpicture}[
  font=\footnotesize, >=Stealth, node distance=4.5mm,
  box/.style={draw, rounded corners, align=center, inner sep=3pt,
              minimum height=12mm, text width=1.8cm},
  artifact/.style={box, fill=black!4}]
\node[box] (workload) {\textbf{Task + fault}\\ configuration};
\node[box, right=of workload] (execution) {\textbf{Instrumented}\\ agentic execution};
\node[artifact, right=of execution] (raw) {\textbf{Raw trace}\\ Langfuse JSON};
\node[artifact, right=of raw] (sanitized) {\textbf{Sanitized trace}\\ \texttt{case.json}};
\node[artifact, right=of sanitized] (structured) {\textbf{Structured view}\\ one line per span};
\node[box, right=of structured] (detector) {\textbf{Diagnosis method}\\ fault type + location};
\node[box, right=of detector] (score) {\textbf{Score}\\ prediction vs. held-out label};

\draw[->] (workload) -- (execution);
\draw[->] (execution) -- (raw);
\draw[->] (raw) -- (sanitized);
\draw[->] (sanitized) -- (structured);
\draw[->] (structured) -- (detector);
\draw[->] (sanitized.north) -- ++(0,3mm)
  -| node[pos=0.25, above, font=\scriptsize\bfseries] {Raw view} (detector.north);
\draw[->] (detector) -- (score);
\end{tikzpicture}
\caption{The \benchmark{} data-generation and diagnosis pipeline.}
\Description{A task and fault configuration drives an instrumented agentic
execution, producing a raw Langfuse trace. Ground-truth fields are removed to
produce a sanitized trace, which is used directly or summarized into a structured
view for diagnosis. Predictions are scored against separately held-out labels.}
\label{fig:overview}
\end{figure*}
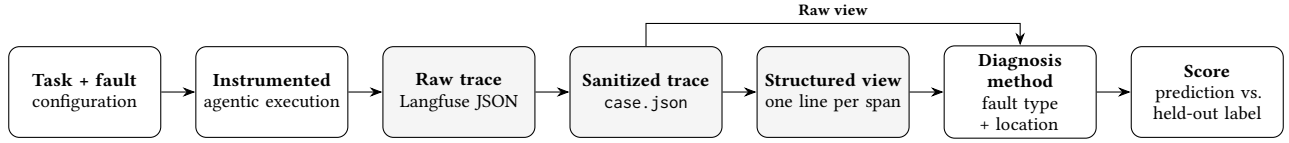

Production-like agent executions expose a broad fault surface. A remote tool may become unavailable, time out, or return a plausible but corrupted response; an LLM call may exceed its context budget; and an inter-agent request may be delayed or routed to the wrong agent. These faults are not necessarily caused by an incorrect reasoning step, yet they can propagate through a long execution and eventually appear as an incorrect answer, excessive latency, or a stalled workflow. Initial work has therefore adapted fault injection and chaos-engineering ideas to agentic systems, studying tool unavailability, latency, hangs, incorrect responses, task perturbations, and tool or API failures~\cite{iannillo2025chaosllm,gupta2026reliabilitybench}. More broadly, a recent benchmark of real-world agent-system maintenance found that state-of-the-art software-engineering agents resolved only $0.67$--$4.67\%$ of the studied issues~\cite{rahardja2025agentissues}, underscoring how difficult agent-specific failures remain to repair. These efforts establish the importance of testing and maintaining agents under stress, but provide limited support for diagnosing faults across heterogeneous tool, model, guardrail, and inter-agent boundaries.

In parallel, failure-attribution research has shown that diagnosing agentic systems is itself difficult. Existing studies catalog recurring behavioral failure modes in multi-agent systems and use techniques such as counterfactual replay and programmatic perturbation to identify failure-inducing agents and steps in long trajectories~\cite{cemri2025mast,zhang2025agentracer}. This line of work primarily attributes failures to agent actions or reasoning behavior~\cite{deshpande2025trail,epperson2025interactive,wang2026flat,zhu2025llm,Anna2026tracelevel}. It does not directly address the complementary operational question considered in this paper: given telemetry from an agentic execution, can a detector distinguish normal behavior from an externally induced runtime fault, identify the fault type, and localize the affected system component? Answering this question requires executions with controlled faults, system-level telemetry, and ground truth that is independent of a detector's interpretation of the final answer.

We introduce \benchmark, a fault-injection benchmark for evaluating the diagnosability of LLM-based agentic systems.\footnote{The experimental code and a sample of the benchmark data are available at \url{https://github.com/kevinzck8k/agentic-fault-diagnosis}.} We use agent applications derived from AI-NativeBench~\cite{wang2026ainativebench} as executable workloads and inject faults at multiple boundaries of their execution, including tool calls, LLM calls, guardrails, and agent-to-agent interactions. Each run is executed with a locally deployed LLM and instrumented through Langfuse to capture its distributed trace, including agent, model, and tool spans and their timing, inputs, outputs, and status metadata. Fault-injection markers and labels are removed before diagnosis; the injected fault type and affected location are retained separately as ground truth.

The resulting benchmark spans five heterogeneous agentic systems and ten fault types (including tool failure, excessive tool or inter-agent latency, context overflow, tool output corruption, tool and agent misrouting, and guardrail bypass) plus a no-fault control, for $275$ controlled executions with same-input alignment between faulty and fault-free runs. The benchmark formulates diagnosis as two related tasks: fault-type classification and localization of the responsible span or component. Because explicit failures and subtle semantic corruptions coexist in the same taxonomy, the benchmark tests both direct recognition of error signals and contextual reasoning over an execution trace. Its held-out labels and compact prediction format also support reproducible comparison of LLM-based and non-LLM diagnosis methods. A first set of LLM baselines shows the task is far from solved: local detectors up to $14$B parameters stay at $13.6$--$19.2\%$ top-1 accuracy over the ten fault types, and even the frontier DeepSeek-v4-pro reaches only $24.8\%$. Component localization also remains unreliable: top-1 accuracy reaches only $31\%$, and getting type and location jointly right stays at $8$--$22\%$. The reference-dependent faults most relevant to trustworthiness (above all a bypassed guardrail) remain near-unsolved from a single trace even for the frontier model. An aligned fault-free reference raises context-overflow recall by up to $55$ points and helps selected latency and routing faults, but its effect is fault-specific and a bypassed guardrail remains unresolved.

In summary, this paper makes the following contributions:
\begin{itemize}[leftmargin=*]
    \item We define and implement a production-oriented fault model spanning
    tool, model, guardrail, and inter-agent boundaries.
    \item We construct $275$ sanitized traces across five systems and ten
    fault types plus no-fault controls, with aligned references and verified type
    and location labels.
    \item We provide an automated diagnosis protocol and LLM baselines showing
    low fault-type and localization accuracy and fault-specific gains from paired
    references.
\end{itemize}

\section{Benchmark Design and Construction}
\label{sec:method}

An agentic system is not a single model, but a composition of probabilistic
language models, orchestration logic, local and remote tools, guardrails, and
networked services. Modern systems increasingly connect these components through
the Model Context Protocol (MCP) for tool access and the Agent-to-Agent (A2A)
protocol for inter-agent communication~\cite{wang2026ainativebench}. This
composition makes reliability a property of the complete execution rather than
of the model or final output alone. Assuring agentic systems therefore requires
examining \emph{how} an execution unfolded in addition to \emph{what} it produced.

\benchmark{} supports this form of analysis through controlled fault injection
and trace-based diagnosis. We execute five heterogeneous agentic systems, inject
faults into their model calls, tools, guardrails, and inter-agent interactions,
and capture the resulting telemetry as structured traces. A diagnosis method
receives a trace with all injection markers and fault labels removed, and must
classify the fault and localize the affected span or component. To make this task
both realistic and verifiable, injected faults must model plausible operational
failures and produce evidence in the collected telemetry that grounds their
assigned fault type and location. Figure~\ref{fig:overview} summarizes the
resulting pipeline.

\subsection{Design Goals}

The benchmark design follows six goals:
\begin{itemize}[leftmargin=*]
    \item \textbf{Production-relevant faults.} Faults model concrete operational
    failures (unreachable or slow tools, corrupted responses, context-budget
    exhaustion, delayed or misrouted delegations, bypassed guardrails) rather
    than synthetic reasoning perturbations.
    \item \textbf{Coverage across system components and interactions.} The fault
    taxonomy spans LLM calls, tools, guardrails, individual agents, and
    inter-agent communication across multiple agent frameworks.
    \item \textbf{Reproducibility.} Every case is produced by a fixed set of
    task inputs executed against a locally deployed model under a controlled
    injection configuration, so results do not depend on a proprietary API whose
    behavior may drift.
    \item \textbf{No ground-truth leakage.} Injection markers, fault labels, and
    other attributes that directly reveal the injected fault are removed from the
    diagnosis input; a method receives only telemetry available through the
    observability pipeline.
    \item \textbf{Trace-grounded labels.} Each case has a held-out fault type and
    affected span or component. Quality-control checks verify that the
    injected fault produces a distinguishing signal in the trace relative to an
    aligned fault-free run of the same input (\S\ref{sec:qc}).
    \item \textbf{Automated scoring.} A machine-readable prediction format
    supports scoring LLM and non-LLM methods without human adjudication.
\end{itemize}

\begin{table*}[t]
\centering
\caption{Selected AI-NativeBench~\cite{wang2026ainativebench} workloads used in
\benchmark{}.}
\label{tab:systems}
\small
\begin{tabular}{@{}p{3.0cm}p{5.2cm}ccp{4.8cm}@{}}
\toprule
\textbf{System} & \textbf{Workflow} & \textbf{\# Agents} & \textbf{\# Tools} &
\textbf{Agent-framework mapping} \\
\midrule
SQL Assistant & Generates SQL from natural language, checks compliance with a
revision loop, executes the query, and interprets the result & 4 & 8 & \mbox{CrewAI}:
generator, reviewer; \mbox{LangGraph}: checker; \mbox{AutoGen}: interpreter \\
Book Writer & Researches and outlines a book, drafts chapters concurrently, and
validates the consolidated manuscript & 5 & 9 & \mbox{AutoGen}: researcher, planner;
\mbox{CrewAI}: chapter researcher, drafter; \mbox{LangGraph}: QA lead \\
Social Media Manager & Analyzes a topic, generates a stylized post, and repeats
generation when a verifier rejects it & 3 & 7 & \mbox{LangGraph}: topic analyst; \mbox{CrewAI}:
content generator; \mbox{AutoGen}: post verifier \\
Landing Page Generator & Expands a product idea, selects a template, and creates
and validates the final HTML page & 3 & 4 & \mbox{LangGraph}: idea analyst; \mbox{AutoGen}:
template selector; \mbox{CrewAI}: content editor \\
Recruitment Assistant & Analyzes job requirements, evaluates candidates, and
prepares interview and communication materials & 3 & 5 & \mbox{LangGraph}: job analyst;
\mbox{CrewAI}: candidate evaluator; \mbox{AutoGen}: interview coordinator \\
\bottomrule
\end{tabular}
\end{table*}

\subsection{Agentic Systems and Task Selection}
\label{sec:systems}

We build \benchmark{} on five agentic applications from
AI-NativeBench~\cite{wang2026ainativebench}, selected so that the benchmark
exercises the boundaries our fault taxonomy targets. Concretely, we required each
selected system to (i) use real tool or MCP interaction, (ii) delegate work
across agents over the A2A protocol, (iii) produce multi-step executions long
enough for faults to propagate, and (iv) admit a deterministic success condition.
We use the heterogeneous A2A (H-A2A) variant of each system. In this variant,
agents implemented with CrewAI, LangGraph, and AutoGen communicate through A2A
and access external tools through MCP. Because this protocol stack is common to
all five systems, Table~\ref{tab:systems} instead shows how agent roles are
assigned to frameworks in each workflow.

\subsection{Deployment and Execution Environment}
\label{sec:deploy}

Each system is deployed as a set of independent A2A services (one process per
participating agent) exposing HTTP/JSON-RPC endpoints, together with the tool and
MCP servers the workflow requires. An orchestrator issues a task input and drives
the workflow to completion or failure.

To make executions reproducible and independent of any external LLM provider, all
agents are served by a single locally deployed model, Qwen3.5-9B, hosted with the
vLLM inference server behind an OpenAI-compatible endpoint. All agents in a run
share this endpoint, so behavior differences between the fault-free and
fault-injected runs of the same input are attributable to the injected fault
rather than to model variation. Decoding is greedy (temperature $0$), and tool
calling uses vLLM's Hermes-style parser. Bounded timeouts and retries are applied
at the A2A and tool boundaries, matching how such systems behave in deployment;
these policies are themselves part of the observed behavior a detector must
reason about. For example, bounded retries make repeated attempts visible as
recurring calls in the trace.

\subsection{Fault Model}
\label{sec:faultmodel}

Our fault model is intended to cover major operational failure modes rather than
serve as an exhaustive ontology. It spans four ways the environment can perturb
an agentic execution: \emph{availability failures} make a tool or A2A invocation
fail explicitly (\faultname{Tool Failure}, \faultname{A2A Timeout});
\emph{performance faults} delay an otherwise valid interaction
(\faultname{Tool Latency}, \faultname{A2A Latency}); \emph{control and routing faults}
alter what is invoked, repeated, or delegated (\faultname{Infinite Loop},
\faultname{Tool Misroute}, \faultname{Agent Misroute}); and \emph{data and policy
faults} alter exchanged content or enforcement decisions
(\faultname{Context Overflow}, \faultname{Output Corruption}, \faultname{Guardrail
Bypass}). Together, these classes exercise tool, LLM, agent, inter-agent, and
guardrail boundaries. They describe the injected mechanism, not its expected
diagnosis difficulty.

Table~\ref{tab:faults} defines the ten fault types together with the boundary
they target, the observable signal they are designed to leave in the trace, and
the intended ground-truth location. A no-fault control accompanies every input.
For readability, we use spaced display names in the paper (e.g.,
\faultname{Tool Failure}); released labels use the corresponding snake\_case
identifiers (e.g., \texttt{tool\_failure}).
We distinguish an \emph{injected runtime fault} from an
\emph{intrinsic reasoning error}: the benchmark injects faults into the execution
environment (tools, delegations, guardrails, model I/O) while holding the model
and task input fixed. For every input, we first collect a \emph{clean} execution
with no injection active; each fault-injected execution then reuses that input,
yielding an aligned clean execution that serves as a fault-free reference. This pairing establishes that the
diagnosed fault reflects an environmental condition rather than a model mistake,
and provides the basis for our quality control (\S\ref{sec:qc}).

\begin{table*}[t]
\centering
\caption{Fault types, target boundaries, observable trace signals, and expected
diagnosis difficulty, grouped by the four fault-mechanism classes of
\S\ref{sec:faultmodel}.}
\label{tab:faults}
\small
\begin{tabular}{@{}llp{4.2cm}p{4.7cm}p{2.6cm}@{}}
\toprule
\textbf{Fault type} & \textbf{Boundary} & \textbf{Description} &
\textbf{Observable signal} & \textbf{Expected difficulty} \\
\midrule
\multicolumn{5}{@{}l}{\emph{Availability failures}} \\
\faultname{Tool Failure} & Tool & The invoked tool returns an error instead
of a result. & The affected tool span records an error status. & Low (explicit
error) \\
\faultname{A2A Timeout} & Inter-agent & An inter-agent delegation exceeds its
timeout. & The affected A2A span records a timeout error. & Low (explicit error) \\
\midrule
\multicolumn{5}{@{}l}{\emph{Performance faults}} \\
\faultname{Tool Latency} & Tool & A tool response is delayed by approximately
15 seconds. & The tool span includes a 15~s delay. & Low (duration) \\
\faultname{A2A Latency} & Inter-agent & An inter-agent response is delayed by
approximately 15 seconds. & The A2A span includes the delay and variable
downstream work. & Medium (duration, noisy) \\
\midrule
\multicolumn{5}{@{}l}{\emph{Control and routing faults}} \\
\faultname{Infinite Loop} & Tool & A bounded retry forces the agent to repeat
the same tool call. & At least three calls share the same tool name and input. &
Low (repeated calls) \\
\faultname{Tool Misroute} & Tool & A call intended for one tool is redirected
to a different tool. & The called tool differs from the clean run and returns
incompatible data. & High (expected tool) \\
\faultname{Agent Misroute} & Agent & A task is delegated to an agent other than
its intended recipient. & The destination differs from the clean run. &
Moderate (expected recipient) \\
\midrule
\multicolumn{5}{@{}l}{\emph{Data and policy faults}} \\
\faultname{Context Overflow} & LLM & An oversized response is added to the
model's context. & The output exceeds $1.3\times$ the clean output size. &
Moderate (clean comparison) \\
\faultname{Output Corruption} & Tool & The injector modifies the content of a
tool response. & The tool output differs from the clean run. & High
(clean comparison) \\
\faultname{Guardrail Bypass} & Guardrail & A request that the guardrail should
block is allowed to proceed. & The guardrail records \emph{pass} for a
policy-violating request. & High (policy judgment) \\
\bottomrule
\end{tabular}
\end{table*}

\subsection{Fault-Injection Mechanism}
\label{sec:injection}

Faults are introduced by interceptors placed at the boundaries between an agent
and the resource it invokes (its tools, its model calls, its guardrail checks,
and its A2A delegations) so that injection is transparent to agent logic and
identical in placement across the five systems. An injection configuration names
a target boundary, an activation condition (which input, which call index), and
an operator. The operators mirror the fault families of
Table~\ref{tab:faults}: delaying a response, raising an exception, replacing a
payload with a corrupted or oversized one, forcing a bounded retry to repeat an
identical call, overriding a guardrail verdict, or redirecting a delegation to a
different agent endpoint.

Two properties of the mechanism are essential to the benchmark's validity. First,
injection is \emph{argument-isolated} where a fault could otherwise perturb the
model: for control and routing faults such as \faultname{Infinite Loop} we drive the
repeated call through a dedicated anchor argument signature, so the loop is
observable as repeated identical calls without destabilizing the surrounding
generation. Second, each fault is realized against a \emph{genuine} target: for
\faultname{Tool Misroute}, the misrouted call must reach a tool whose behavior truly
differs from the intended one, so that the fault produces a real output mismatch
rather than a no-op. When a fault fires, the injector records the
activation (the fault type, the affected span, and its location) into a
separate ground-truth record. This record is used to construct
\texttt{labels.jsonl} and is never written into the trace the detector sees.

\subsection{Observability and Trace Collection}
\label{sec:observability}

A useful diagnosis trace must expose more than LLM and tool calls: it must also
reveal agent identity, guardrail decisions, and A2A interactions with per-step
timing and status~\cite{balusu2026agenttelemetry}. We instrument every run with Langfuse,
which records a hierarchical trace of spans covering the execution. Each span
carries an identifier, a span kind, a level/status, start time and duration,
and, subject to filtering, its input and output. Observed span kinds include
\texttt{AGENT} (an agent's activity), \texttt{LLM\_CALL}, \texttt{TOOL\_CALL},
\texttt{MEMORY}, and the A2A delegation spans that connect agents, alongside
structural \texttt{CHAIN}/\texttt{SPAN} nodes; guardrail decisions surface as
their own spans. A typical execution comprises on the order of tens to a few
hundred spans (e.g., $64$ spans for a representative SQLAssistant run).

Before release, traces are normalized and filtered: injection markers and any
fields that would leak the label are removed, and the directory path that
organizes a case by fault type (used only for storage) is explicitly \emph{not}
part of the detector input. What remains is the span hierarchy with timing,
status, identity, and content that a production observability stack would expose.

\subsection{Dataset Construction and Specification}

\stitle{Case Generation and Quality Control}
\label{sec:qc}
Cases are generated by crossing the five systems with the ten fault types plus a
no-fault control, over a fixed set of five task inputs per condition, with inputs
\emph{aligned} across conditions so that every faulty case has a same-input
fault-free reference. This yields $5\times 11\times 5 = 275$ cases ($55$ per
system: $50$ faulty and $5$ no-fault).

Because the controlled design fixes the input set, we do not discard or resample
cases to obtain clean-looking numbers; instead every case must be made valid by
ensuring the intended fault actually fired and left its signal. We enforce this
with an automated audit that, for each faulty case, compares the trace against
its aligned clean execution and checks that the fault's expected signal
(Table~\ref{tab:faults}) is present, for example, that \faultname{Infinite Loop}
produces at least three calls with the same tool name and input, that
\faultname{Context Overflow} produces an output more than 1.3 times the size of
the corresponding clean output, and that \faultname{Guardrail Bypass} records a
\emph{pass} decision for a policy-violating request. Cases that did not exhibit their signal
were traced to the underlying injection mechanism and fixed at the source rather
than masked. For the availability, control/routing, and data/policy faults this
yields an unambiguous signal in every released case. The two performance faults
differ. \faultname{Tool Latency} is clean: the injected ${\sim}15$\,s lands on a
short tool span, and its measured delta is $+15.0$\,s at the median. \faultname{A2A
Latency} is inherently softer: the same delay is added to a delegation span that
already brackets the downstream agent's variable work, so the measured delta ranges
from ${-}3$\,s to ${+}26$\,s across systems; the injection fires, but its signal is
often masked. We surface this difficulty rather than mask it, and it is reflected
in the low diagnosis rates for \faultname{A2A Latency}.

\stitle{Trace Representation}
\label{sec:representation}
Each released case \texttt{case\_NN.json} contains the sanitized trace provided
to the detector: a list of spans, each with an identifier, name, span
kind, level, start time, duration, and filtered input/output. Because real
multi-agent traces are large (in our data the full-trace serialization has a
median of ${\sim}108$K tokens and reaches ${\sim}638$K tokens for the largest
case), feeding the raw trace to a detector is often impractical. We therefore also
define a compact \emph{structured view} that renders one line per span exposing
the quantitative features a diagnosis needs: span id, kind, name, level,
duration, output size, a repeat count of identical (name+input) calls, and a short
output preview (Figure~\ref{fig:structured}). The structured view reduces the
largest traces by roughly an order of magnitude while
surfacing (not computing) standard observability features; it deliberately does
not flag anomalies or compare against a baseline, as that would perform the
diagnosis for the detector. For the single outlier, a BookWriter
\faultname{Context Overflow} run containing $1{,}591$ identical framework
event-queue polls, we collapse identical (name+input) spans into one row while
retaining their true repeat count in the \texttt{xN} column, bringing the view
within the local detectors' context windows without changing the other $274$
views.

\begin{figure}[t]
\centering
\footnotesize
\begin{SaveVerbatim}{RawSpan}
{"id":"85f8974d..","name":"Crew..kickoff",
 "duration_ms":132501,
 "input":"{chapter_title:..,goal:..}",
 "output":"{title: The Nature and Power of
   Illusions, content: # ...}" (12,012 chars)}
\end{SaveVerbatim}
\begin{SaveVerbatim}{StructuredSpan}
[85f8974d] span Crew..kickoff | DEFAULT |
  132501 | 12012 | x1 | {title: The Nature
  and Power of Illusions, content: # ...
\end{SaveVerbatim}
\begin{minipage}{0.86\columnwidth}
\textbf{Raw span description}
\textrm{(13{,}802 JSON characters; ${\sim}100$ spans per trace)}
\par\smallskip
\begingroup
\setlength{\fboxsep}{4pt}
\colorbox{black!6}{%
  \begin{minipage}{\dimexpr\linewidth-2\fboxsep\relax}
  \UseVerbatim{RawSpan}
  \end{minipage}}
\endgroup

\medskip
\textbf{Structured span description}
\textrm{(one 245-character line)}
\par\smallskip
\begingroup
\setlength{\fboxsep}{4pt}
\colorbox{black!6}{%
  \begin{minipage}{\dimexpr\linewidth-2\fboxsep\relax}
  \UseVerbatim{StructuredSpan}
  \end{minipage}}
\endgroup
\end{minipage}
\caption{Raw and structured representations of one trace span.}
\Description{The raw representation shows a gray box containing multiline JSON.
The structured representation shows a smaller gray box containing one line with
the span identifier, kind, name, level, duration, output size, repeat count, and
output preview.}
\label{fig:structured}
\end{figure}

\stitle{Ground Truth and Dataset Summary}
\label{sec:groundtruth}
Ground-truth answers are held out in \texttt{labels.jsonl}, keyed by case UID.
Each record gives the fault type and a numeric fault id, the required span kind,
the detection signal, and the affected location (span kind, component name, and
call index); for routing faults it additionally records the misrouted target.
Table~\ref{tab:faults} lists the per-fault detection signal and required kind. The
benchmark and evaluation kit, including the audit and scoring scripts, will be
released publicly.

\section{Evaluation}
\label{sec:eval}

We ask whether a diagnosis method can infer a fault's presence, type, and
responsible component from sanitized telemetry, and how that depends on the fault
class and detector capacity. We report a first set of LLM baselines, organized
around four research questions:

\begin{itemize}[leftmargin=*]
    \item \textbf{RQ1 (Fault-type diagnosis and trace representation).} How
    accurately can a detector identify the fault type from a single
    sanitized trace, and how do the structured and raw views affect coverage
    and accuracy?
    \item \textbf{RQ2 (Fault-specific difficulty).} How does diagnosability vary
    across fault types and the four fault-model classes?
    \item \textbf{RQ3 (Localization).} Beyond naming the fault type, can a detector
    reliably identify the responsible component and jointly predict its location
    and fault type?
    \item \textbf{RQ4 (Reference condition).} Does providing a comparable
    fault-free reference trace improve diagnosis over the single-trace setting,
    and for which fault types?
\end{itemize}

\subsection{Diagnosis Task and Prediction Format}

In the primary single-trace setting, a method receives one execution trace of
unknown status (which may be fault-free) and must return a ranked list of
candidate fault types drawn from the eleven conditions (the ten faults plus
\faultname{No Fault}), each with a short textual justification citing spans. From
the ranking we score top-1 and top-3 correctness. The optional paired-trace
setting (RQ4) additionally supplies a known-normal execution of the same input;
because inputs and external task state are held fixed across conditions
(\S\ref{sec:deploy}), the two traces differ only in the injected fault.

\subsection{Baselines}

Our baselines are general-purpose instruction-tuned LLMs applied zero-shot. To span
an order of magnitude in detector capacity we use four open models (Qwen3-1.7B,
Qwen3-4B, Qwen3.5-9B, and Qwen3-14B), each served locally
with vLLM, and we add the frontier DeepSeek-v4-pro (served via API) as a strong
upper reference. Every detector is run under both trace representations where its
context window permits (\S\ref{sec:representation}), so the same model is compared
on the structured and raw views. All detectors share an identical prompt that asks
for a ranked list of candidate fault types with span-citing justifications. A
rule-based detector over the observable signals in Table~\ref{tab:faults} is a
natural non-LLM baseline left to future work.

\subsection{Metrics}

For fault-type diagnosis we report top-1 accuracy (AC@1) and top-3 accuracy
(AC@3) over the faulty cases, and separately the rate at which no-fault controls
are correctly identified as \faultname{No Fault} (a proxy for the false-positive
behavior of a detector). We report accuracy overall and per fault type; because
the eleven conditions are balanced by construction, a uniform-random detector
scores $1/11\approx 9\%$ AC@1. For localization we report, over the $225$ cases
whose label carries a location (the $25$ \faultname{Context Overflow} faults and $25$
\faultname{No Fault} controls have none), component-location AC@1 and AC@3: the
top-1, respectively any top-3, cited span must resolve to the ground-truth
component after normalizing span-kind and delegation-call name decorations.
Type-and-location AC@1 requires both the top-ranked fault type and the predicted
component to be correct.

\subsection{Experimental Setup}

All baselines run against the corrected \benchmark{} dataset of $275$ cases, with
directory names and held-out fields excluded from the input and greedy decoding
throughout. On the structured view every detector covers all $275$ cases, so all
cross-detector comparisons use identical samples. Outputs that yield no parseable
ranking are counted as incorrect, and their rate varies sharply by detector:
$0\%$ for DeepSeek-v4-pro and $2\%$ for Qwen3-14B, but $5\%$, $17\%$, and $33\%$
for Qwen3-1.7B, Qwen3-4B, and Qwen3.5-9B. These are truncations (finish reason
\emph{length}) in which a reasoning model exhausts its output budget
mid-deliberation; we show in \S\ref{sec:discussion} that doubling the budget leaves
the accuracies unchanged, so they are not an artifact of the budget. The raw view
is only attempted where the context window
allows: it is infeasible for the ${\le}40960$-token Qwen3 detectors, Qwen3.5-9B
runs raw within its $262$K native window (covering the $225/275$ cases that fit),
and DeepSeek-v4-pro within its $1$M window (all $275$). Scoring is fully automated
against \texttt{labels.jsonl} by case UID. DeepSeek-v4-pro's localization scores
come from a separate API run whose fault-type accuracy
($25.2\%/37.6\%$ AC@1/AC@3) differs from Table~\ref{tab:overall} by less than four
percentage points.

\subsection{Results}
\label{sec:results}

Table~\ref{tab:overall} reports AC@1 and AC@3 on the $250$ faulty cases for the
structured and raw trace representations; a dash indicates that the detector
could not ingest the raw trace within its context window. Table~\ref{tab:perfault}
reports per-fault-type top-3 recall on structured traces. Each entry is the number
of cases whose true fault type appears in the detector's top three, out of $25$
cases ($5$ systems $\times$ $5$ inputs); the final row instead counts no-fault
controls correctly recognized as clean.

\begin{table}[t]
\centering
\caption{Fault-type diagnosis accuracy.}
\label{tab:overall}
\small
\begin{tabular}{@{}lcccc@{}}
\toprule
\textbf{Trace format} & \multicolumn{2}{c}{\textbf{Structured view}} &
\multicolumn{2}{c}{\textbf{Raw trace}} \\
\cmidrule(lr){2-3}\cmidrule(lr){4-5}
\textbf{Detector} & AC@1 & AC@3 & AC@1 & AC@3 \\
\midrule
Qwen3-1.7B        & 15.6\% & 24.0\% & --     & --     \\
Qwen3-4B          & 13.6\% & 31.2\% & --     & --     \\
Qwen3.5-9B        & 19.2\% & 28.0\% & 15.2\% & 22.5\% \\
Qwen3-14B         & 17.2\% & \textbf{35.2\%} & --     & --     \\
DeepSeek-v4-pro   & \textbf{24.8\%} & 34.0\% & \textbf{21.6\%} & \textbf{36.0\%} \\
\bottomrule
\end{tabular}
\end{table}

\begin{table}[t]
\centering
\caption{Per-fault-type top-3 recall on structured traces.}
\label{tab:perfault}
\small
\setlength{\tabcolsep}{4.5pt}
\begin{tabular}{@{}lccccc@{}}
\toprule
\textbf{Model family} & \multicolumn{4}{c}{\textbf{Qwen}} &
\multicolumn{1}{c}{\textbf{DeepSeek}} \\
\cmidrule(lr){2-5}\cmidrule(l){6-6}
\textbf{Fault type} & \textbf{3-1.7B} & \textbf{3-4B} &
\textbf{3.5-9B} & \textbf{3-14B} & \textbf{v4-pro} \\
\midrule
\multicolumn{6}{@{}l}{\emph{Availability failures}} \\
\faultname{Tool Failure}      & 24/25 & 24/25 & 23/25 & \textbf{25/25} & 24/25 \\
\faultname{A2A Timeout}       &  6/25 & 20/25 & 19/25 & 20/25 & \textbf{23/25} \\
\midrule
\multicolumn{6}{@{}l}{\emph{Performance faults}} \\
\faultname{A2A Latency}       & 10/25 & 10/25 & 12/25 & \textbf{13/25} &  7/25 \\
\faultname{Tool Latency}      &  0/25 &  1/25 &  7/25 &  9/25 & \textbf{11/25} \\
\midrule
\multicolumn{6}{@{}l}{\emph{Control and routing faults}} \\
\faultname{Infinite Loop}     &  9/25 & \textbf{12/25} &  5/25 &  6/25 &  2/25 \\
\faultname{Tool Misroute}     &  0/25 &  4/25 &  0/25 & \textbf{5/25} &  3/25 \\
\faultname{Agent Misroute}    &  0/25 &  1/25 &  2/25 &  3/25 & \textbf{10/25} \\
\midrule
\multicolumn{6}{@{}l}{\emph{Data and policy faults}} \\
\faultname{Context Overflow}  & \textbf{8/25} &  2/25 &  2/25 &  1/25 &  0/25 \\
\faultname{Output Corruption} &  3/25 &  3/25 &  0/25 & \textbf{5/25} &  4/25 \\
\faultname{Guardrail Bypass}  &  0/25 & \textbf{1/25} &  0/25 & \textbf{1/25} & \textbf{1/25} \\
\midrule
\faultname{No Fault} (recognized) &  1/25 & 14/25 & 12/25 & \textbf{24/25} & 22/25 \\
\bottomrule
\end{tabular}
\end{table}

\stitle{RQ1: fault-type diagnosis remains difficult across trace representations}
Table~\ref{tab:overall} shows that fault-type diagnosis from a single
sanitized trace is difficult. The local Qwen detectors sit at
$13.6$--$19.2\%$ AC@1, barely above the $9\%$ random baseline and essentially flat
across an order of magnitude of model size; and even DeepSeek-v4-pro, a frontier
model, reaches only $24.8\%$ AC@1 ($34.0\%$ AC@3). Larger models do get the correct fault into their
top-3 more often, but for most cases they cannot commit to it as the first choice.
Inspecting the $14$B detector, the true fault is absent from the top-3 entirely in
$65\%$ of faulty cases, so the AC@1--AC@3 gap reflects genuine misses rather than
mere ranking
noise, and only $2\%$ of outputs are malformed: this is a capability gap, not a
formatting artifact. Recognition of no-fault controls follows a different trend:
it rises from $4\%$ for Qwen3-1.7B to $96\%$ for Qwen3-14B (and $88\%$ for
DeepSeek-v4-pro), indicating that scale helps detectors avoid false alarms even
though fault-type classification remains weak.

Trace representation also determines whether diagnosis is computationally
feasible. The local detectors with context windows of at most $131$K tokens
cannot ingest raw traces that reach ${\sim}638$K tokens. Qwen3.5-9B can process
the $82\%$ of cases that fit its $262$K window, but its raw-view accuracy is lower
than its structured-view accuracy ($15.2\%/22.5\%$ versus
$19.2\%/28.0\%$ AC@1/AC@3), and $40\%$ of its raw-view outputs contain no
parseable ranking. DeepSeek-v4-pro can ingest all raw traces; its raw and
structured results are similar ($21.6\%/36.0\%$ versus $24.8\%/34.0\%$), while
the raw run costs approximately $8.6\times$ more (\$16.45 versus \$1.91 for the
$275$ cases) because it contains ${\sim}14\times$ as many prompt tokens. The
structured view therefore expands detector coverage and reduces inference cost
without increasing the strongest detector's accuracy artificially.

\stitle{RQ2: mechanism class does not by itself determine diagnosability}
Table~\ref{tab:perfault} follows the fault taxonomy of
Table~\ref{tab:faults}. Availability failures are the easiest overall:
\faultname{Tool Failure} reaches $25/25$ and \faultname{A2A Timeout} $23/25$.
Performance faults and \faultname{Infinite Loop} carry directly
observable signals (a large duration or repeated call) but detectors use them
unevenly ($0/25$ to $13/25$), showing that visible evidence does not guarantee
correct classification. The other control/routing faults require knowledge of the
intended destination: local detectors remain near chance on \faultname{Tool
Misroute} and \faultname{Agent Misroute}, although the frontier model raises the
latter to $40\%$. Data and policy faults are similarly difficult because their
traces can remain well formed. In particular, \faultname{Guardrail Bypass} is
usually predicted as \faultname{No Fault}: a forged \emph{pass} verdict is
indistinguishable from a legitimate approval without knowing that the request
should have been rejected. The smallest detector's apparent edge on
\faultname{Context Overflow} ($8/25$) is a calibration artifact rather than skill: it
over-predicts that condition on unrelated cases, buying recall with a matching
false-positive rate while almost never recognizing a clean run (no-fault recognition
itself scales sharply, from $1/25$ at $1.7$B to $24/25$ at $14$B). Thus, across the
four mechanism classes, the more predictive divide is whether the evidence can be
interpreted directly or requires a reference for the \emph{expected} behavior.

\begin{table}[t]
\centering
\caption{Component-location and type-and-location accuracy on structured traces.}
\label{tab:localization}
\small
\begin{tabular}{@{}lccc@{}}
\toprule
\textbf{Detector} & \multicolumn{2}{c}{\textbf{Location}} &
\multicolumn{1}{c}{\textbf{Type + location}} \\
\cmidrule(lr){2-3}\cmidrule(l){4-4}
& \textbf{AC@1} & \textbf{AC@3} & \textbf{AC@1} \\
\midrule
Qwen3-1.7B      & 23\% & 27\% &  8\% \\
Qwen3-4B        & 20\% & 30\% &  9\% \\
Qwen3.5-9B      & 27\% & 36\% & 16\% \\
Qwen3-14B       & 27\% & 44\% & 15\% \\
DeepSeek-v4-pro & \textbf{31\%} & \textbf{45\%} & \textbf{22\%} \\
\bottomrule
\end{tabular}
\end{table}

\stitle{RQ3: reliable localization remains difficult}
Table~\ref{tab:localization} scores whether a detector points to the responsible
component, over the $225$ faulty cases that carry a location. Top-1 localization
remains between $20\%$ and $31\%$, so even the frontier detector identifies the
wrong component in more than two thirds of cases. Allowing three candidates
raises localization to $27\%$--$45\%$, but no detector localizes a majority of
cases. Type-and-location AC@1 is lower still: $8\%$--$16\%$ for the local
detectors and $22\%$
for DeepSeek-v4-pro. Scale improves location AC@3 more consistently than
fault-type AC@1, but the absolute results show that both component localization
and type-and-location diagnosis remain open problems.

\begin{figure}[htbp]
\centering
\begin{tikzpicture}
\begin{axis}[
  width=1.02\columnwidth, height=5.4cm,
  ybar, bar width=3.6pt,
  ymin=-26, ymax=62,
  ylabel={$\Delta$ top-3 recall (percentage points)},
  ylabel style={font=\scriptsize}, ylabel near ticks,
  yticklabel style={font=\scriptsize},
  symbolic x coords={Context Overflow,Agent Misroute,Tool Latency,Infinite Loop,
                     Tool Misroute,A2A Latency,A2A Timeout,Tool Failure,
                     Output Corruption,Guardrail Bypass},
  xtick=data,
  x tick label style={rotate=50, anchor=east, font=\scriptsize},
  legend style={at={(0.5,1.03)}, anchor=south, legend columns=3,
                draw=none, font=\scriptsize, column sep=4pt},
  ymajorgrids, grid style={dotted, gray!50},
  enlarge x limits=0.06,
]
\fill[green!8]  (axis description cs:0,0.295) rectangle (axis description cs:1,1);
\fill[red!7]    (axis description cs:0,0)     rectangle (axis description cs:1,0.295);
\addplot+[bar shift=-3.6pt] coordinates {(Context Overflow,10) (Agent Misroute,4) (Tool Latency,0)
  (Infinite Loop,26) (Tool Misroute,11) (A2A Latency,30) (A2A Timeout,-20)
  (Tool Failure,0) (Output Corruption,-15) (Guardrail Bypass,0)};
\addplot+[bar shift=0pt] coordinates {(Context Overflow,5) (Agent Misroute,40) (Tool Latency,0)
  (Infinite Loop,0) (Tool Misroute,12) (A2A Latency,-5) (A2A Timeout,4)
  (Tool Failure,0) (Output Corruption,-10) (Guardrail Bypass,-5)};
\addplot+[bar shift=3.6pt] coordinates {(Context Overflow,55) (Agent Misroute,16) (Tool Latency,35)
  (Infinite Loop,11) (Tool Misroute,0) (A2A Latency,-5) (A2A Timeout,0)
  (Tool Failure,0) (Output Corruption,-5) (Guardrail Bypass,0)};
\legend{Qwen3-1.7B, Qwen3-4B, Qwen3-14B}
\draw[black!55, semithick] (axis description cs:0,0.295) -- (axis description cs:1,0.295);
\node[anchor=north east, align=right, font=\scriptsize\itshape, text=green!45!black]
  at (axis description cs:0.99,0.88) {higher recall with\\paired traces $\uparrow$};
\node[anchor=south west, align=left, font=\scriptsize\itshape, text=red!60!black]
  at (axis description cs:0.01,0.03) {$\downarrow$ higher recall with\\a single trace};
\end{axis}
\end{tikzpicture}
\caption{Change in top-3 recall after adding a fault-free reference trace (relative to single-trace diagnosis).}
\Description{A grouped bar chart showing the change in top-3 recall between paired
and single-trace diagnosis for ten fault types and three Qwen detectors. Positive
bars indicate higher recall with paired traces; negative bars indicate higher recall
with a single trace.}
\label{fig:rq4}
\end{figure}
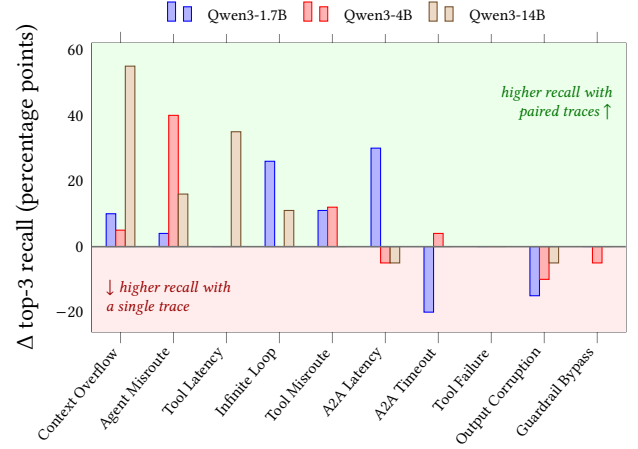

\stitle{RQ4: aligned references improve selected fault types}
We test the reference condition directly: each faulty trace is paired with the
aligned \faultname{No Fault} run of the same input, given to the detector as a
known-normal baseline (\S\ref{sec:representation}). Figure~\ref{fig:rq4} reports the
change in top-3 recall, $\Delta =$ paired $-$ single, on matched cases. The reference
helps when $\Delta$ is positive. It lifts overall AC@3 for every detector, and
the lift grows with capability
($+3.9$, $+4.9$, and $+10.7$ percentage points for $1.7$B, $4$B, and $14$B).
The largest gains occur for faults defined by a deviation from normal behavior:
for Qwen3-14B, the reference raises \faultname{Context Overflow} recall by $55$
points and \faultname{Tool Latency} by $35$ points, while routing faults improve
for some detectors. These changes show that an aligned execution supplies useful
comparison signals for output size, duration, and destination, but do not imply a
general solution to single-trace diagnosis.

Two exceptions sharpen the picture. \faultname{Guardrail Bypass} is unmoved
or slightly worse ($-5$ to $0$ points): on a benign input the reference guard
also returns \emph{pass}, so a forged pass is indistinguishable even side by side.
\faultname{Output Corruption} also decreases by $5$--$15$ points because a
plausible but incorrect output does not expose a consistent textual difference
that the detectors can interpret as corruption. Using a reference also demands
capacity: Qwen3.5-9B reasons past its output budget on the doubled
input and fails to emit a valid ranking on over half of paired cases even at $8192$
output tokens, so we omit it. Overall, paired traces improve several faults with
measurable baselines, but they do not help faults that require an external policy
or semantic judgment.

\stitle{Implications}
Two findings bear on assurance. A diagnosis pipeline must first solve a
representation problem, since raw traces reach hundreds of thousands of tokens
(\S\ref{sec:representation}). And the faults that matter most for trustworthiness,
above all a bypassed guardrail and a silent context overflow, are exactly the ones
single-trace detectors miss at every scale, motivating the reference-based and
specialized methods the benchmark is designed to evaluate.

\section{Related Work}
\label{sec:related}

\stitle{Agentic systems and observability}
Benchmarks such as AgentBench established the breadth of tasks LLM agents can
solve through multi-step environment interaction~\cite{liu2024agentbench}, and
AI-native suites like AI-NativeBench extend this to white-box systems that combine
agents with MCP tools and A2A communication~\cite{wang2026ainativebench}. As these
systems grow into distributed software, operating them reliably requires telemetry
beyond final-answer checking. AgentTelemetry argues that useful agent
observability must expose agent identity, guardrails, delegation, and
timing/status metadata, and provides a fault-detection toolkit over such
telemetry~\cite{balusu2026agenttelemetry}. \benchmark{} adopts this observability
stance and pairs it with controlled fault injection and held-out ground truth.

\stitle{Fault injection and stress testing for agents}
Chaos-engineering ideas have begun to reach agents~\cite{basiri2016chaos}. ChaosLLM injects unreachable,
slow, hanging, and incorrect responses at the tool boundary of a ReAct
agent~\cite{iannillo2025chaosllm}, and ReliabilityBench evaluates consistency,
robustness to task perturbations, and tolerance of tool/API faults under
production-like stress~\cite{gupta2026reliabilitybench}. These works establish the
value of stressing agents but focus on single-agent tool faults and on outcome
robustness rather than on \emph{diagnosing} a fault across heterogeneous
boundaries. \benchmark{} injects faults at the tool, LLM, guardrail, agent, and
inter-agent boundaries of multi-framework, multi-agent systems, and evaluates
diagnosis from traces stripped of injection metadata and labels.

\stitle{Failure attribution in multi-agent systems}
A parallel line studies who or what caused a failure. MAST catalogs recurring
behavioral failure modes of multi-agent systems~\cite{cemri2025mast}; AgenTracer
uses counterfactual replay and programmatic perturbation to attribute failures to
agents and steps in long trajectories~\cite{zhang2025agentracer}; spectrum-based
methods adapt fault-localization statistics to agent
executions~\cite{ge2025famas}; and recent work rethinks attribution from
multiple perspectives~\cite{in2026rethinking}. LLMGuard targets fault diagnosis
for language-model services~\cite{zhong2026llmguard}. This body of work primarily
attributes failures to \emph{internal} agent actions or reasoning. \benchmark{}
is complementary: it targets \emph{external}, environmentally induced runtime
faults, with ground truth that is independent of the final answer and of any
detector's interpretation, and it explicitly includes a no-fault control so that
false positives are measurable.

\section{Discussion}
\label{sec:discussion}

\subsection{Limitations}

The current benchmark uses five systems with five aligned inputs per condition;
while this yields $275$ verified cases and balanced per-fault coverage, broader
task and input diversity would strengthen external validity, and we plan to scale
both. Our baselines are zero-shot general-purpose LLMs (four local Qwen models and the
frontier DeepSeek-v4-pro) evaluated on both the structured and raw trace views; a
rule-based non-LLM detector, few-shot prompting, and specialized attribution
methods remain to be evaluated. Because DeepSeek-v4-pro is served through a
commercial API, its outputs are not reproducible even at temperature $0$: a re-run
disagrees on nearly half ($48\%$) of top-1 predictions, though aggregate accuracy is stable to
within a point, so we read its per-case results as a single sample while the local
Qwen detectors are fully reproducible. We report fault-type diagnosis and
component-level localization; exact-span
scoring and richer localization metrics (e.g., MRR) remain to be added. Finally, all runs use a single locally served model to hold
generation fixed across conditions, which aids reproducibility but leaves
cross-model generalization of the \emph{workloads} (as opposed to the detectors)
for future study.

\subsection{Threats to Validity}

\stitle{Internal validity}
Four factors could distort our scores; we address each. First, an injected fault
must be real and its identity hidden: we
remove injection markers and label-bearing fields from the released trace and
verify that every faulty case exhibits its intended signal against an aligned
no-fault reference (\S\ref{sec:qc}), fixing cases that failed this check at the
injection source rather than dropping them, which would bias the set toward easy
cases. Second, because faulty and fault-free runs share the same input and model
endpoint, observed trace differences are attributable to the injected fault.
Third, trace representation is a potential confound, but the structured view only
surfaces observability features without computing anomaly judgments, and the one
model able to read both views scores the same on each
(Table~\ref{tab:overall}), so the compact view does not inflate accuracy, while the
paired-trace experiment (RQ4, Figure~\ref{fig:rq4}) shows that the benefit of a
reference is concentrated in specific fault types rather than uniformly improving
diagnosis. Fourth, the Qwen reasoning
detectors are sometimes truncated before emitting a ranking (up to $33\%$ for
Qwen3.5-9B, $\leq 2\%$ for Qwen3-14B and DeepSeek-v4-pro); doubling the output budget
($4096\to 8192$ tokens) leaves both truncation ($37\%\to 33\%$ for Qwen3.5-9B) and
accuracy (its AC@1 $18.4\to 19.2$) essentially unchanged, so these are hard cases on
which deliberation diverges rather than answers a larger budget would recover, and
scoring them incorrect does not understate ability.

\stitle{External validity}
The systems, faults, and single-trace protocol
approximate but do not exhaust production conditions; real deployments mix faults,
vary load, and evolve over time. The taxonomy targets common operational failures
at standard agent boundaries, and the injection architecture is framework-agnostic
by placement, which we expect to transfer, but confirming this on additional
systems is future work.

\section{Conclusion and Future Work}
\label{sec:conclusion}

We presented \benchmark, a benchmark of $275$ sanitized traces for
classifying and localizing ten runtime fault types across five multi-agent
systems. Its controlled injections, aligned no-fault executions, verified labels,
and automated scoring support reproducible evaluation of agent-system diagnosis.

The evaluated LLMs remain unreliable: single-trace fault-type AC@1 reaches at
most $24.8\%$, component-location AC@1 at most $31\%$, and type-and-location AC@1
at most $22\%$. A compact structured view makes long traces tractable without
improving the strongest detector artificially. Aligned references improve context overflow (by up to $55$ points) and selected
latency and routing faults, but not faults requiring semantic or policy judgment,
including guardrail bypass. These results
motivate reference-based and specialized detectors, non-LLM baselines, and
exact-span localization, all of which the released benchmark is designed to
support.

\bibliographystyle{ACM-Reference-Format}
\bibliography{references}


\begin{thebibliography}{19}


\ifx \showCODEN    \undefined \def \showCODEN     #1{\unskip}     \fi
\ifx \showISBNx    \undefined \def \showISBNx     #1{\unskip}     \fi
\ifx \showISBNxiii \undefined \def \showISBNxiii  #1{\unskip}     \fi
\ifx \showISSN     \undefined \def \showISSN      #1{\unskip}     \fi
\ifx \showLCCN     \undefined \def \showLCCN      #1{\unskip}     \fi
\ifx \shownote     \undefined \def \shownote      #1{#1}          \fi
\ifx \showarticletitle \undefined \def \showarticletitle #1{#1}   \fi
\ifx \showURL      \undefined \def \showURL       {\relax}        \fi
\providecommand\bibfield[2]{#2}
\providecommand\bibinfo[2]{#2}
\providecommand\natexlab[1]{#1}
\providecommand\showeprint[2][]{arXiv:#2}

\bibitem[Balusu(2026)]%
        {balusu2026agenttelemetry}
\bibfield{author}{\bibinfo{person}{Krishna~Chaitanya Balusu}.} \bibinfo{year}{2026}\natexlab{}.
\newblock \showarticletitle{{AgentTelemetry}: A Fault Detection Benchmark and Toolkit for {LLM} Agent Observability}. In \bibinfo{booktitle}{\emph{Proceedings of the 3rd ACM International Conference on AI-Powered Software}}. \bibinfo{pages}{380--387}.
\newblock


\bibitem[Basiri et~al\mbox{.}(2016)]%
        {basiri2016chaos}
\bibfield{author}{\bibinfo{person}{Ali Basiri}, \bibinfo{person}{Niosha Behnam}, \bibinfo{person}{Ruud De~Rooij}, \bibinfo{person}{Lorin Hochstein}, \bibinfo{person}{Luke Kosewski}, \bibinfo{person}{Justin Reynolds}, {and} \bibinfo{person}{Casey Rosenthal}.} \bibinfo{year}{2016}\natexlab{}.
\newblock \showarticletitle{Chaos Engineering}.
\newblock \bibinfo{journal}{\emph{IEEE Software}} \bibinfo{volume}{33}, \bibinfo{number}{3} (\bibinfo{year}{2016}), \bibinfo{pages}{35--41}.
\newblock


\bibitem[Cemri et~al\mbox{.}(2025)]%
        {cemri2025mast}
\bibfield{author}{\bibinfo{person}{Mert Cemri}, \bibinfo{person}{Melissa~Z Pan}, \bibinfo{person}{Shuyi Yang}, \bibinfo{person}{Lakshya~A Agrawal}, \bibinfo{person}{Bhavya Chopra}, \bibinfo{person}{Rishabh Tiwari}, \bibinfo{person}{Kurt Keutzer}, \bibinfo{person}{Aditya Parameswaran}, \bibinfo{person}{Dan Klein}, \bibinfo{person}{Kannan Ramchandran}, {et~al\mbox{.}}} \bibinfo{year}{2025}\natexlab{}.
\newblock \showarticletitle{Why Do Multi-Agent {LLM} Systems Fail?}. In \bibinfo{booktitle}{\emph{Advances in Neural Information Processing Systems}}.
\newblock


\bibitem[Deshpande et~al\mbox{.}(2025)]%
        {deshpande2025trail}
\bibfield{author}{\bibinfo{person}{Darshan Deshpande}, \bibinfo{person}{Varun Gangal}, \bibinfo{person}{Hersh Mehta}, \bibinfo{person}{Jitin Krishnan}, \bibinfo{person}{Anand Kannappan}, {and} \bibinfo{person}{Rebecca Qian}.} \bibinfo{year}{2025}\natexlab{}.
\newblock \showarticletitle{{TRAIL}: Trace Reasoning and Agentic Issue Localization}.
\newblock \bibinfo{journal}{\emph{arXiv preprint arXiv:2505.08638}} (\bibinfo{year}{2025}).
\newblock


\bibitem[Epperson et~al\mbox{.}(2025)]%
        {epperson2025interactive}
\bibfield{author}{\bibinfo{person}{Will Epperson}, \bibinfo{person}{Gagan Bansal}, \bibinfo{person}{Victor~C Dibia}, \bibinfo{person}{Adam Fourney}, \bibinfo{person}{Jack Gerrits}, \bibinfo{person}{Erkang Zhu}, {and} \bibinfo{person}{Saleema Amershi}.} \bibinfo{year}{2025}\natexlab{}.
\newblock \showarticletitle{Interactive Debugging and Steering of Multi-Agent {AI} Systems}. In \bibinfo{booktitle}{\emph{Proceedings of the 2025 CHI Conference on Human Factors in Computing Systems}}.
\newblock


\bibitem[Ge et~al\mbox{.}(2025)]%
        {ge2025famas}
\bibfield{author}{\bibinfo{person}{Yu Ge}, \bibinfo{person}{Linna Xie}, \bibinfo{person}{Zhong Li}, \bibinfo{person}{Yu Pei}, {and} \bibinfo{person}{Tian Zhang}.} \bibinfo{year}{2025}\natexlab{}.
\newblock \showarticletitle{Who Is Introducing the Failure? Automatically Attributing Failures of Multi-Agent Systems via Spectrum Analysis}.
\newblock \bibinfo{journal}{\emph{arXiv preprint arXiv:2509.13782}} (\bibinfo{year}{2025}).
\newblock


\bibitem[Gupta(2026)]%
        {gupta2026reliabilitybench}
\bibfield{author}{\bibinfo{person}{Aayush Gupta}.} \bibinfo{year}{2026}\natexlab{}.
\newblock \showarticletitle{{ReliabilityBench}: Evaluating {LLM} Agent Reliability Under Production-Like Stress Conditions}.
\newblock \bibinfo{journal}{\emph{arXiv preprint arXiv:2601.06112}} (\bibinfo{year}{2026}).
\newblock


\bibitem[He et~al\mbox{.}(2025)]%
        {he2025llm}
\bibfield{author}{\bibinfo{person}{Junda He}, \bibinfo{person}{Christoph Treude}, {and} \bibinfo{person}{David Lo}.} \bibinfo{year}{2025}\natexlab{}.
\newblock \showarticletitle{{LLM}-Based Multi-Agent Systems for Software Engineering: Literature Review, Vision, and the Road Ahead}.
\newblock \bibinfo{journal}{\emph{ACM Transactions on Software Engineering and Methodology}} (\bibinfo{year}{2025}).
\newblock


\bibitem[Iannillo(2025)]%
        {iannillo2025chaosllm}
\bibfield{author}{\bibinfo{person}{Antonio~Ken Iannillo}.} \bibinfo{year}{2025}\natexlab{}.
\newblock \showarticletitle{{ChaosLLM}: A Dependability Testing Approach for Tool-Calling Agents}. In \bibinfo{booktitle}{\emph{2025 IEEE 36th International Symposium on Software Reliability Engineering Workshops (ISSREW)}}.
\newblock


\bibitem[In et~al\mbox{.}(2026)]%
        {in2026rethinking}
\bibfield{author}{\bibinfo{person}{Yeonjun In}, \bibinfo{person}{Mehrab Tanjim}, \bibinfo{person}{Jayakumar Subramanian}, \bibinfo{person}{Sungchul Kim}, \bibinfo{person}{Uttaran Bhattacharya}, \bibinfo{person}{Wonjoong Kim}, \bibinfo{person}{Sangwu Park}, \bibinfo{person}{Somdeb Sarkhel}, {and} \bibinfo{person}{Chanyoung Park}.} \bibinfo{year}{2026}\natexlab{}.
\newblock \showarticletitle{Rethinking Failure Attribution in Multi-Agent Systems: A Multi-Perspective Benchmark and Evaluation}.
\newblock \bibinfo{journal}{\emph{arXiv preprint arXiv:2603.25001}} (\bibinfo{year}{2026}).
\newblock


\bibitem[Liu et~al\mbox{.}(2024)]%
        {liu2024agentbench}
\bibfield{author}{\bibinfo{person}{Xiao Liu}, \bibinfo{person}{Hao Yu}, \bibinfo{person}{Hanchen Zhang}, \bibinfo{person}{Yifan Xu}, \bibinfo{person}{Xuanyu Lei}, \bibinfo{person}{Hanyu Lai}, \bibinfo{person}{Yu Gu}, \bibinfo{person}{Hangliang Ding}, \bibinfo{person}{Kaiwen Men}, \bibinfo{person}{Kejuan Yang}, {et~al\mbox{.}}} \bibinfo{year}{2024}\natexlab{}.
\newblock \showarticletitle{{AgentBench}: Evaluating {LLMs} as Agents}. In \bibinfo{booktitle}{\emph{International Conference on Learning Representations}}.
\newblock


\bibitem[Mazhar et~al\mbox{.}(2026)]%
        {Anna2026tracelevel}
\bibfield{author}{\bibinfo{person}{Anna Mazhar}, \bibinfo{person}{Huzaifa Suri}, {and} \bibinfo{person}{Sainyam Galhotra}.} \bibinfo{year}{2026}\natexlab{}.
\newblock \showarticletitle{Trace-Level Analysis of Information Contamination in Multi-Agent Systems}. In \bibinfo{booktitle}{\emph{Proceedings of the ACM Conference on AI and Agentic Systems}}.
\newblock


\bibitem[Rahardja et~al\mbox{.}(2025)]%
        {rahardja2025agentissues}
\bibfield{author}{\bibinfo{person}{Alfin~Wijaya Rahardja}, \bibinfo{person}{Junwei Liu}, \bibinfo{person}{Weitong Chen}, \bibinfo{person}{Zhenpeng Chen}, {and} \bibinfo{person}{Yiling Lou}.} \bibinfo{year}{2025}\natexlab{}.
\newblock \showarticletitle{Can Agents Fix Agent Issues?}. In \bibinfo{booktitle}{\emph{NeurIPS}}.
\newblock


\bibitem[Wang et~al\mbox{.}(2026a)]%
        {wang2026flat}
\bibfield{author}{\bibinfo{person}{Yawen Wang}, \bibinfo{person}{Wenjie Wu}, \bibinfo{person}{Junjie Wang}, {and} \bibinfo{person}{Qing Wang}.} \bibinfo{year}{2026}\natexlab{a}.
\newblock \showarticletitle{From Flat Logs to Causal Graphs: Hierarchical Failure Attribution for {LLM}-Based Multi-Agent Systems}.
\newblock \bibinfo{journal}{\emph{arXiv preprint arXiv:2602.23701}} (\bibinfo{year}{2026}).
\newblock


\bibitem[Wang et~al\mbox{.}(2026b)]%
        {wang2026ainativebench}
\bibfield{author}{\bibinfo{person}{Zirui Wang}, \bibinfo{person}{Guangba Yu}, {and} \bibinfo{person}{Michael~R Lyu}.} \bibinfo{year}{2026}\natexlab{b}.
\newblock \showarticletitle{{AI-NativeBench}: An Open-Source White-Box Agentic Benchmark Suite for {AI}-Native Systems}.
\newblock \bibinfo{journal}{\emph{arXiv preprint arXiv:2601.09393}} (\bibinfo{year}{2026}).
\newblock


\bibitem[Zhang et~al\mbox{.}(2025)]%
        {zhang2025agentracer}
\bibfield{author}{\bibinfo{person}{Guibin Zhang}, \bibinfo{person}{Junhao Wang}, \bibinfo{person}{Junjie Chen}, \bibinfo{person}{Wangchunshu Zhou}, \bibinfo{person}{Kun Wang}, {and} \bibinfo{person}{Shuicheng Yan}.} \bibinfo{year}{2025}\natexlab{}.
\newblock \showarticletitle{{AgenTracer}: Who Is Inducing Failure in the {LLM} Agentic Systems?}
\newblock \bibinfo{journal}{\emph{arXiv preprint arXiv:2509.03312}} (\bibinfo{year}{2025}).
\newblock


\bibitem[Zhao et~al\mbox{.}(2026)]%
        {zhao2026debugging}
\bibfield{author}{\bibinfo{person}{Chenyu Zhao}, \bibinfo{person}{Shenglin Zhang}, \bibinfo{person}{Yihang Lin}, \bibinfo{person}{Wenwei Gu}, \bibinfo{person}{Zhimin Chen}, \bibinfo{person}{Yongqian Sun}, \bibinfo{person}{Dan Pei}, \bibinfo{person}{Chetan Bansal}, \bibinfo{person}{Saravan Rajmohan}, {and} \bibinfo{person}{Minghua Ma}.} \bibinfo{year}{2026}\natexlab{}.
\newblock \showarticletitle{Debugging the Debuggers: Failure-Anchored Structured Recovery for Software Engineering Agents}.
\newblock \bibinfo{journal}{\emph{arXiv preprint arXiv:2605.08717}} (\bibinfo{year}{2026}).
\newblock


\bibitem[Zhong et~al\mbox{.}(2026)]%
        {zhong2026llmguard}
\bibfield{author}{\bibinfo{person}{Yuedong Zhong}, \bibinfo{person}{Guangba Yu}, \bibinfo{person}{Yujie Huang}, \bibinfo{person}{QunChao Fu}, \bibinfo{person}{Rui Ren}, \bibinfo{person}{Cong Feng}, \bibinfo{person}{Yongqiang Yang}, {and} \bibinfo{person}{Michael Lyu}.} \bibinfo{year}{2026}\natexlab{}.
\newblock \showarticletitle{{LLMGuard}: Multi-Agent Fault Diagnosis for Reliable Language-Model-as-a-Service}. In \bibinfo{booktitle}{\emph{2026 56th Annual IEEE International Conference on Dependable Systems and Networks (DSN)}}.
\newblock


\bibitem[Zhu et~al\mbox{.}(2025)]%
        {zhu2025llm}
\bibfield{author}{\bibinfo{person}{Kunlun Zhu}, \bibinfo{person}{Zijia Liu}, \bibinfo{person}{Bingxuan Li}, \bibinfo{person}{Muxin Tian}, \bibinfo{person}{Yingxuan Yang}, \bibinfo{person}{Jiaxun Zhang}, \bibinfo{person}{Pengrui Han}, \bibinfo{person}{Qipeng Xie}, \bibinfo{person}{Fuyang Cui}, \bibinfo{person}{Weijia Zhang}, {et~al\mbox{.}}} \bibinfo{year}{2025}\natexlab{}.
\newblock \showarticletitle{Where {LLM} Agents Fail and How They Can Learn from Failures}.
\newblock \bibinfo{journal}{\emph{arXiv preprint arXiv:2509.25370}} (\bibinfo{year}{2025}).
\newblock


\end{thebibliography}

\end{document}